\pdfoutput=1

\documentclass[11pt]{article}

\usepackage{Transformer_Transferability-arxiv}
\usepackage[T1]{fontenc}
\usepackage[utf8]{inputenc}
\usepackage{times, inconsolata, booktabs, multirow, colortbl, graphicx,bm}
\input{dictionary.cls}

\title{An Empirical Study on the Transferability of Transformer Modules\\in Parameter-Efficient Fine-Tuning}

\author{Mohammad Akbar-Tajari\thanks{$~~$The authors contributed equally to this work.}$~~^1$, Sara Rajaee\samethanks$~~^2$, \and Mohammad Taher Pilehvar$^3$\\
	$^1$ Sharif University of Technology, Iran\\
	$^2$ University of Amsterdam, Netherlands\\
	$^3$ Tehran Institute for Advanced Studies, Iran\\
	\texttt{m.akbarTajari@gmail.com}\quad\texttt{s.rajaee@uva.nl}\quad\texttt{mp792@cam.ac.uk}}

\begin{document}
	\maketitle

	\begin{abstract}
		Parameter-efficient fine-tuning approaches have recently garnered a lot of attention. Having considerably lower number of trainable weights, these methods can bring about scalability and computational effectiveness. In this paper, we look for optimal sub-networks and investigate the capability of different transformer modules in transferring knowledge from a pre-trained model to a downstream task. Our empirical results suggest that \textit{every} transformer module in BERT can act as a \textit{winning ticket}: fine-tuning each specific module while keeping the rest of the network frozen can lead to comparable performance to the full fine-tuning. Among different modules, \layernorms\ exhibit the best capacity for knowledge transfer with limited trainable weights, to the extent that, with only 0.003\% of all parameters in the layer-wise analysis, they show acceptable performance on various target tasks. On the reasons behind their effectiveness, we argue that their notable performance could be attributed to their high-magnitude weights compared to that of the other modules in the pre-trained BERT. The code for this paper is freely available at \url{https://github.com/m-tajari/transformer-transferability}.
	\end{abstract}

	\section{Introduction}
		Fine-tuning is widely used as a procedure to employ the knowledge learned during pre-training of language models for specific tasks \cite{howard-ruder-2018-universal,peters-etal-2019-tune,merchant-etal-2020-happens,zhou-srikumar-2022-closer}. However, fine-tuning can be a computationally expensive process, given that it usually involves updating all the parameters in transformer-based models which are often massive in size. Parameter-efficient fine-tuning methods try to ameliorate this by reducing the number of updatable parameters during fine-tuning.

		Adapters \cite{pmlr-v97-houlsby19a,pfeiffer-etal-2020-adapterhub,wang-etal-2021-k,ruckle-etal-2021-adapterdrop,hu2021lora} try to circumvent this issue by inserting light-weight modules in the transformer blocks, tuning of which usually results in comparable performance to the full fine-tuning (while the number of updatable parameters is significantly lower). Nevertheless, introducing new parameters to an already-large model can be considered a drawback. Another category of parameter-efficient fine-tuning methods is based on the \textit{Lottery Ticket Hypothesis} \cite{prasanna-etal-2020-bert}, where the goal is to find a small subset of parameters that can compete with the full fine-tuning setting. Various subsets of network parameters have been suggested as the \textit{winning ticket}, including the connections with high magnitudes \cite{NIPS2015_ae0eb3ee}, identity mappings \cite{lin-etal-2020-pruning}, and dominant dimensions \cite{guo-etal-2021-parameter}.

		In this paper, we study the ability of different modules of a transformer block in knowledge transfer. Our experiments provide a more comprehensive analysis than the existing work, which usually suggests specific modules as the winning ticket, such as the bias terms \cite{ben-zaken-etal-2022-bitfit}. Through module-wise fine-tuning, we check if the winning ticket is a property that can be associated only with some specific modules in the transformer block. Our results suggest that \textit{all} individual modules possess this property to some extent. Among these, \layernorms\ prove to be highly reliable for knowledge transfer: fine-tuning only $37k$ \layernorm\ weights (out of $110M$ parameters in BERT-base) is often on par with full fine-tuning on various downstream tasks. Extending this analysis, we show that tuning even only one \layernorm\ can yield comparable performance and that the middle layers are the best in terms of transferability. We also investigate the reasons behind the effectiveness of \layernorm\ tuning. Our experiments suggest that this could be due to the relatively high-magnitude weights in these modules. In fact, we show that tuning just a tiny fraction of high-magnitude dimensions (usually referred to as \textit{outliers}) can lead to competitive performance on various tasks.

	\section{Winning Modules}
		According to the Lottery Ticket Hypothesis, there are small sub-networks whose performance is comparable to the over-parameterized model on different tasks \cite{frankle2018the}. Several studies have been carried out to identify sub-networks across the model that can provide the best transferability \cite{Gale2019mag,rigl,lee2021layeradaptive,guo-etal-2021-parameter,hu2021lora}. Nonetheless, finding the winning sub-network usually requires extra computation, which is costly in terms of time and memory. In this section, we take another look at the transformer block of BERT and focus on the ability of its different modules to transfer knowledge to various downstream tasks. More specifically, we aim to find the winning module among the different modules in the transformer-based architecture of the pre-trained BERT.

		\begin{table*}[ht!]
	\centering
	\setlength{\tabcolsep}{4pt}
	\scalebox{0.89}{
		\begin{tabular}{l c c c c c c c c c c c}
			\toprule
			\keyFieldL{Model} & \keyFieldL{\param} & \keyFieldS{\cola} & \keyFieldS{\sst} & \keyFieldS{\mrpc} & \keyFieldS{\sts} & \keyFieldS{\qqp} & \keyFieldS{\mnlim} & \keyFieldS{\mnlimm} & \keyFieldS{\qnli} & \keyFieldS{\rte} & \keyFieldL{\it Avg.}\\
			\midrule
			\keyFieldL{\fullfinetune} & \ent{100.0\%} & \entry{54.7}{0.7} & \bentry{93.1}{0.1} & \entry{90.9}{1.0} & \sbentry{89.0}{0.5} & \bentry{87.3}{0.1} & \bentry{83.4}{0.2} & \bentry{83.7}{0.4} & \bentry{91.5}{0.1} & \entry{71.7}{1.2} & \sbent{82.8}\\
			\midrule
			\keyFieldL{\feedforward} & \ent{51.76\%} & \sbentry{56.2}{1.5} & \entry{92.7}{0.1} & \sbentry{91.1}{0.5} & \entry{88.9}{0.6} & \sbentry{87.1}{0.1} & \entry{81.9}{0.3} & \entry{82.7}{0.2} & \entry{90.8}{0.0} & \entry{71.1}{0.8} & \ent{82.5}\\
			\keyFieldL{\multihead} & \ent{25.89\%} & \bentry{59.0}{1.3} & \sbentry{92.9}{0.3} & \bentry{91.2}{0.3} & \bentry{89.1}{0.5} & \entry{86.7}{0.2} & \sbentry{83.2}{0.2} & \sbentry{83.6}{0.3} & \sbentry{91.2}{0.1} & \sbentry{72.0}{0.2} & \bent{83.2}\\
			\cmidrule(lr){2-11}
			\keyFieldL{\layernorms} & \ent{0.03\%} & \entry{52.8}{2.0} & \entry{92.2}{0.4} & \entry{91.0}{0.2} & \entry{88.4}{0.1} & \entry{81.5}{0.1} & \entry{79.3}{0.2} & \entry{80.7}{0.3} & \entry{89.4}{0.2} & \entry{70.2}{0.4} & \ent{80.6}\\
			\keyFieldL{\layernormsA} & \ent{0.02\%} & \entry{50.8}{0.5} & \entry{91.8}{0.2} & \entry{90.6}{0.5} & \entry{88.3}{0.2} & \entry{80.9}{0.1} & \entry{76.3}{0.2} & \entry{75.9}{0.3} & \entry{88.8}{0.1} & \entry{70.5}{0.9} & \ent{79.3}\\
			\keyFieldL{\layernormsF} & \ent{0.02\%} & \entry{52.9}{1.9} & \entry{91.7}{0.2} & \entry{91.0}{0.2} & \entry{88.4}{0.1} & \entry{81.0}{0.1} & \entry{77.0}{0.1} & \entry{77.2}{0.3} & \entry{88.3}{0.0} & \entry{69.2}{0.7} & \ent{79.6}\\
			\midrule
			\keyFieldL{\bitfit} & \ent{0.12\%} & \entry{53.6}{1.9} & \entry{89.6}{0.2} & \entry{90.5}{0.2} & \entry{88.2}{0.0} & \entry{81.8}{0.3} & $^\dagger$\entry{81.4}{0.2} ~ & $^\dagger$\entry{82.2}{0.2} ~ &  $^\dagger$\entry{90.2}{0.2} ~ & \bentry{72.8}{0.2} & \ent{81.1}\\
			\keyFieldL{\rand} & \ent{0.03\%} & \entry{35.5}{3.0} & \entry{86.4}{0.3} & \entry{85.1}{0.5} & \entry{84.3}{0.0} & \entry{74.8}{0.2} & \entry{64.5}{0.2} & \entry{66.1}{0.3} & \entry{76.4}{0.0} & \entry{60.8}{0.5} & \ent{70.4}\\
			\midrule
			\ent{\freeze} & \ent{0.00\%} & \entry{35.2}{1.6} & \entry{81.5}{0.3} & \entry{80.7}{0.1} & \entry{68.3}{0.1} & \entry{60.1}{0.2} & \entry{44.0}{0.3} & \entry{45.1}{0.2} & \entry{68.8}{0.1} & \entry{58.6}{0.5} & \ent{60.3}\\
			\bottomrule
		\end{tabular}}
	\caption{The performance of BERT on the \glue\ benchmark with different fine-tuning strategies. We report Matthew’s correlation for \cola, F1 score for \mrpc\ and \qqp, Spearman's correlation for \sts, and accuracy for the rest. \keyFieldL{\layernormsA}\ (\keyFieldL{\layernormsF}) stands for the scenario in which only \layernorms\ of {$\bf{A}$}ttention ({$\bf{F}$}eedforward) modules are set to be trainable. The best and the second-best results are highlighted for each task.
	$^\dagger$ Results from \citet{ben-zaken-etal-2022-bitfit}.}
	\label{tab:main}
\end{table*}

		\subsection{Experimental Setup}
			\paragraph{Datasets.}
				We fine-tune our models on the \glue\ benchmark \cite{wang-etal-2018-glue}. We leave out the {\wnli} (the Winograd Schema Challenge) task \cite{levesque2012winograd}), given that BERT's performance on this benchmark is not much better than a random classifier. Instead, we test the models on the Corpus of Linguistic Acceptability  \cite[\cola]{warstadt-etal-2019-neural}, the Stanford Sentiment Treebank  \cite[\sst]{socher-etal-2013-recursive}, the Microsoft Research Paraphrase Corpus \cite[\mrpc]{dolan-brockett-2005-automatically}, the Semantic Textual Similarity \cite[\sts]{cer-etal-2017-semeval}, the Quora Question Pairs \cite[\qqp]{wang-etal-2018-glue}, the Multi-Genre Natural Language Inference Corpus \cite[\mnli]{williams-etal-2018-broad}, the Stanford Question Answering Dataset \cite[\qnli]{rajpurkar-etal-2016-squad}, and the Recognizing Textual Entailment  \cite[\rte]{dagan2005pascal}. All the reported results are obtained on the corresponding development sets.

			\paragraph{Models.}
				We opt for bert-base-uncased, implemented by the HuggingFace library in TensorFlow \cite{wolf-etal-2020-transformers,abadi2015tensorflow}. The maximum sequence length is set to 128. Except for the fully fine-tuned model (\fullfinetune), where we train the models for five epochs, the number of epochs is chosen based on the size of the tasks: 10 epochs for \sst, \qqp, \mnli, and \qnli\ and 20 epochs otherwise. We use the Adam optimizer with an epsilon set to 1e-6, a warmup ratio of 10\%, and a batch size of 16. The only hyperparameter tuning we do is on choosing the learning rate from \{1e-5, 5e-5, 1e-4, 5e-4, 1e-3, 5e-3, 1e-2\} to draw a fair comparison with previous work. We report the average and standard deviation of the results of three models trained with different random seeds. All the models are trained on four NVIDIA Tesla V100S-32G GPUs.

			\paragraph{Module Settings.}
				To find out the potential of transformer modules in transfer learning, we pick similar modules across all layers and fine-tune them while keeping the rest of the network frozen. The aim of this setup is to broaden our insights on the distribution of knowledge across the model and the adaptability of different modules to target tasks.

				In every transformer block, we check for the role played by their \textit{\multihead} attention, \textit{\feedforward} layer, and \textit{\layernorms} in knowledge transfer. Since every transformer block has two \layernorms\ (attention and feedforward), we also consider fine-tuning them separately (\textit{\layernormsA} and \textit{\layernormsF}). We also compare against the replicated results of \bitfit\ \cite{ben-zaken-etal-2022-bitfit}, in which consistent bias terms across the transformer blocks are employed for fine-tuning. To verify if consistency in selecting parameters matters, we also show the results of fine-tuning only a small randomly selected subset of all the parameters with the same size as of the \textit{\layernorms} (\textit{\rand}). In the experiments, the full fine-tuning (\textit{\fullfinetune}) and \textit{\freeze} modes are considered as the upper and lower bounds, respectively.

		\subsection{Results}
			Table \ref{tab:main} shows our experimental results on eight tasks from the \glue\ benchmark.\footnote{As for \bitfit, we were unable to carry out full hyperparameter tuning on three tasks due to the large dataset size and computational constraints. Instead, we report the results as given in the original paper, which is around 5\% better than the best results we obtained for these settings.} For each setting, we also report the corresponding ratio of updatable parameters (compared to the full fine-tuning). As can be observed, individual modules of BERT can be considered as \textit{winning tickets} because they can achieve comparable performance to the \textit{\fullfinetune} setting, despite involving significantly smaller numbers of trainable parameters. In particular, \layernorms\ prove to have a high potential in transferability and adaptability to various downstream tasks with a very limited set of trainable parameters ($0.034\%$). The performance is mostly preserved even when only one of the two \layernorms\ is set to be trainable, reducing the number of effective parameters to $0.017\%$ of that in the full fine-tuning. Moreover, our results also reveal that selecting consistent weights (similar modules across layers) has a key role in fine-tuning quality, given that the random subset of a comparable number of parameters does not lead to the same performance levels.

		\subsection{Token-level Classification}
			In addition to the sentence-level tasks of the \glue\ benchmark, we also conduct experiments on two different token-level datasets to broaden our insights on the capacity of individual modules: \penn\ Part-of-speech tagging \cite{marcus-etal-1993-building} and \conl\ Named Entity Recognition \cite{tjong-kim-sang-de-meulder-2003-introduction}. For part-of-speech tagging, we use the subset of the Wall Street Journal (WSJ) portion of PTB which is freely available in the Natural Language Toolkit \cite[NLTK]{bird2009natural}. In this experiment, we adhere to the convention of using the cased version of BERT, given the case-sensitive nature of these token-level tasks.

			Table \ref{tab:token} summarizes the results. Similarly to what is observed on the sentence-level tasks, LeyerNorms can attain competitive performance on both token-level tasks, despite involving just a small fraction of all the model parameters. Moreover, in comparison with the equal number of randomly selected weights, they demonstrate remarkably better performance.

			\begin{table}
	\centering
	\setlength{\tabcolsep}{6.8pt}
	\scalebox{0.9}{
		\begin{tabular}{l c c c}
			\toprule
			\keyFieldL{Model} & \keyFieldL{\param} & \keyFieldL{PTB} & \keyFieldL{CoNLL}\\
			\cmidrule(r){1-1}
			\cmidrule(r){2-2}
			\cmidrule(r){3-3}
			\cmidrule(r){4-4}
			\keyFieldL{\fullfinetune} & \ent{100.0\%} & \bentry{98.5}{0.1} & \bentry{94.2}{0.1} / \bentry{98.6}{0.0}\\
			\keyFieldL{\bitfit} & \ent{0.12\%} & \entry{98.1}{0.1} & \entry{86.9}{0.1} / \entry{96.9}{0.0}\\
			\keyFieldL{\layernorms} & \ent{0.03\%} & \sbentry{98.4}{0.1} & \sbentry{92.4}{0.1} / \sbentry{98.1}{0.0}\\
			\keyFieldL{\rand} & \ent{0.03\%} & \entry{97.6}{0.1} & \entry{82.7}{0.1} / \entry{96.1}{0.0}\\
			\ent{\freeze} & \ent{0.00\%} & \entry{96.2}{0.0} & \entry{78.1}{0.0} / \entry{94.8}{0.0}\\
			\bottomrule
		\end{tabular}}
	\caption{The performance of BERT on \penn\ (PTB) and \conl\ (CoNLL) datasets with five different fine-tuning strategies. We report accuracy for PTB and F1 score (macro/micro) for CoNLL.}
	\label{tab:token}
\end{table}

		\begin{table*}
	\centering
	\setlength{\tabcolsep}{4pt}
	\scalebox{0.84}{
		\begin{tabular}{l c c c c c c c c c c c c c}
			\toprule
			\keyFieldL{Task} & \keyFieldL{\#1} & \keyFieldL{\#2} & \keyFieldL{\#3} & \keyFieldL{\#4} & \keyFieldL{\#5} & \keyFieldL{\#6} & \keyFieldL{\#7} & \keyFieldL{\#8} & \keyFieldL{\#9} & \keyFieldL{\#10} & \keyFieldL{\#11} & \keyFieldL{\#12} & \keyFieldL{\fullfinetune}\\
			\cmidrule(lr){2-13}
			\keyFieldL{\cola} & \entry{44.0}{1.5} & \entry{43.3}{2.8} & \entry{46.8}{2.4} & \sbentry{47.6}{1.7} & \entry{46.1}{1.9} & \entry{47.0}{2.1} & \entry{47.0}{3.4} & \bentry{48.1}{0.9} & \entry{47.1}{2.5} & \entry{45.7}{1.5} & \entry{42.3}{1.9} & \entry{36.9}{0.2} & \bf\entry{54.7}{0.7}\\
			\keyFieldL{\mrpc} & \entry{84.7}{0.5} & \entry{85.9}{0.8} & \entry{84.2}{0.3} & \entry{85.9}{0.6} & \bentry{89.0}{1.1} & \sbentry{88.8}{0.4} & \entry{87.5}{0.7} & \entry{86.1}{0.2} & \entry{86.6}{0.2} & \entry{86.4}{0.7} & \entry{85.2}{0.5} & \entry{82.6}{0.1} & \bf\entry{90.9}{1.0}\\
			\keyFieldL{\sts} & \entry{85.1}{0.3} & \entry{85.7}{0.1} & \entry{86.1}{0.1} & \entry{86.1}{0.2} & \entry{86.7}{0.3} & \sbentry{87.1}{0.1} & \entry{86.9}{0.1} & \bentry{87.2}{0.1} & \entry{86.7}{0.1} & \entry{86.6}{0.1} & \entry{86.5}{0.1} & \entry{83.5}{0.2} & \bf\entry{89.0}{0.5}\\
			\keyFieldL{\rte} & \entry{61.5}{1.7} & \entry{65.3}{0.6} & \entry{63.9}{1.5} & \entry{64.1}{0.6} & \entry{65.2}{0.6} & \entry{64.1}{1.5} & \sbentry{67.3}{0.3} & \bentry{67.5}{1.0} & \entry{63.5}{0.8} & \entry{65.8}{0.3} & \entry{67.0}{0.2} & \entry{60.5}{1.4} & \bf\entry{71.7}{1.2}\\
			\bottomrule
		\end{tabular}}
	\caption{The performance of layer-wise fine-tuning of \layernorms\ on the selected downstream tasks for BERT. The \layernorms\ in the middle layers tend to have the highest transferability.}
	\label{tab:layer}
\end{table*}

		\subsection{Single Norms Tuning}\label{sec:sing}
			Previous studies have reported that different layers do not contribute equally to the ultimate performance in transfer learning \cite{zhou-srikumar-2021-directprobe,rogers-etal-2020-primer,kovaleva-etal-2019-revealing}. We are interested in studying the extent to which individual \layernorms\ in different transformer blocks are adaptable to downstream tasks. To this end, we perform a layer-wise analysis in which the only trainable parameters are the two \layernorms\ in each block and the final classifier. Therefore, the total number of fine-tuning parameters is less than 5K (3,072 and 1,538 for the \layernorms\ and classifier, respectively)\footnote{{For \sts, the number of classifier parameters is 769.}}, which is about $0.003\%$ of all the parameters. Due to our limited computational resources, we restrict our experiments to \cola, \mrpc, \sts, and \rte.

			Table \ref{tab:layer} presents the results for the layer-wise analysis. According to the fine-tuning results, tuning only a single \layernorm\ may be sufficient to achieve performance comparable to fine-tuning all \layernorms. Furthermore, the middle-layer \layernorms\ exhibit the best results across all layers, which can be attributed to the high transferability of the middle layers in BERT, corroborating previous findings on the concentration of task-specific features in these subsets of the network \cite{liu-etal-2019-linguistic}.

	\begin{figure*}[ht!]
		\centering
		\includegraphics[width=157mm, height=63mm]{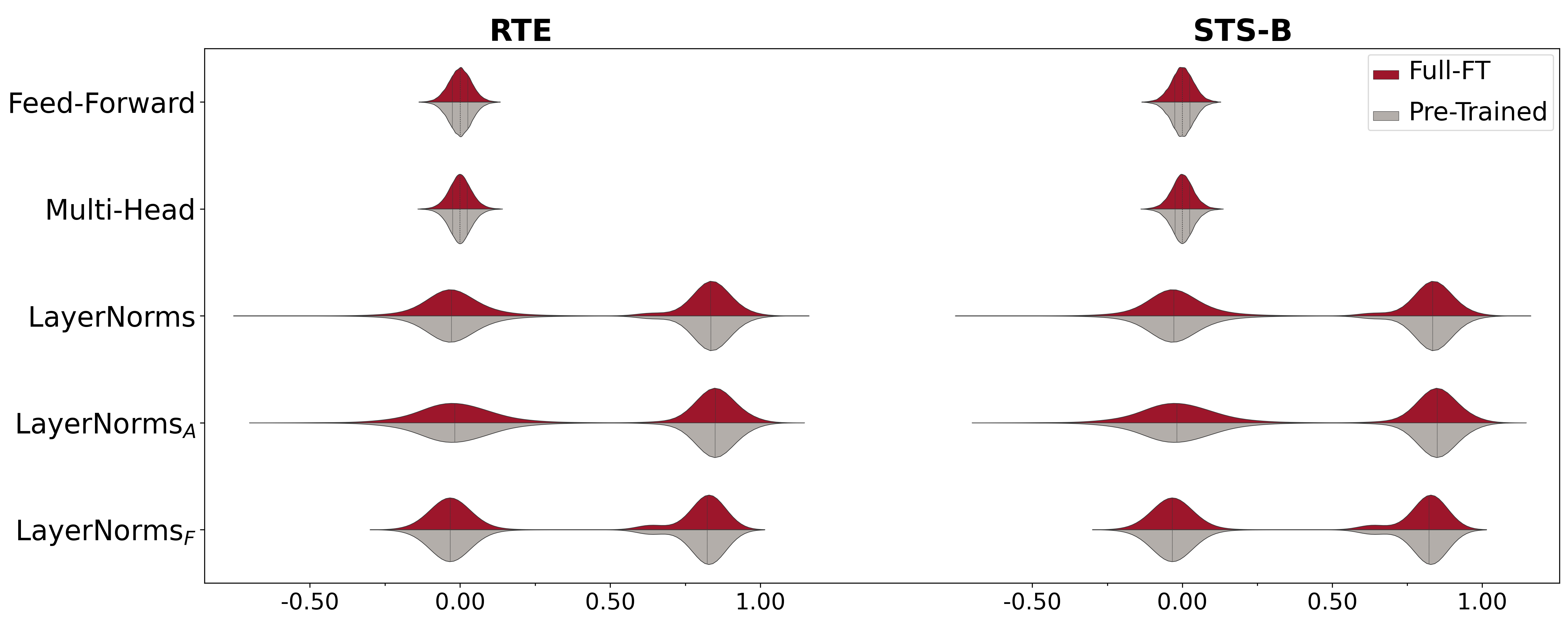}
		\caption{The empirical distribution of a fixed random subset of weights in different modules of BERT. For better visualization, we have discarded the outliers. The weights of \layernorms\ appear to have a bimodal distribution with significantly higher overall averages and standard deviations.}
		\label{fig:analysis}
	\end{figure*}

	\section{Analysis}
		In the previous section, we have shown that different modules of a transformer block can play as winning tickets, since they all have the potential for transferring knowledge to the selected downstream tasks. Among different modules, \layernorms\ have proven to be the most reliable in fine-tuning. In this section, we search for the reasons behind the effectiveness of these modules. To this end, we focus on the magnitude of every weight and how they change during the full fine-tuning across all layers.

		As a first step, in Figure \ref{fig:analysis}, we visualize the distribution of weights for different BERT modules on \rte\ and \sts. In general, the distribution of weights is similar across \feedforward\ and \multihead\ modules. Nevertheless, \layernorms\ tend to have a bimodal distribution, with one of the modes having significantly higher magnitudes\footnote{Notice that in the pre-training process, weights ($\bm\gamma$) and biases ($\bm\beta$) of \layernorms\ were initialized with $\bm1$s and $\bm0$s, respectively, bringing about a distribution consisting of two modes with different averages and standard deviations.}. The pattern is consistent across \layernormsA\ and \layernormsF. We hypothesize that these high-magnitude weights are the reason behind the effectiveness of \layernorms\ and, in what follows, check our hypothesis by restricting our experiments to only high-magnitude dimensions of \layernorms.

		\subsection{Outlier Tuning}\label{sec:outlier}
			\begin{table*}
	\centering
	\setlength{\tabcolsep}{6.8pt}
	\scalebox{0.9}{
		\begin{tabular}{c c c c c c c c c c c c c c c}
			\toprule
			\multirow{2}{*}{\keyFieldL{$n$}} && \multirow{2}{*}{\keyFieldL{\param}} && \multicolumn{2}{c}{\keyFieldL{\cola}} && \multicolumn{2}{c}{\keyFieldL{\mrpc}} && \multicolumn{2}{c}{\keyFieldL{\sts}} && \multicolumn{2}{c}{\keyFieldL{\rte}}\\
			\cmidrule(r){5-6}\cmidrule(r){8-9}\cmidrule(r){11-12}\cmidrule(r){14-15}
			&&&&\keyFieldL{\out}&\keyFieldL{\rand}&&\keyFieldL{\out}&\keyFieldL{\rand}&&\keyFieldL{\out}&\keyFieldL{\rand}&&\keyFieldL{\out}&\keyFieldL{\rand}\\
			\midrule
			\keyFieldL{256} && \ent{0.0056\%} && \entry{55.9}{0.5} & \entry{54.3}{1.6} && \entry{90.3}{0.7} & \entry{89.4}{0.8} && \entry{88.1}{0.1} & \entry{87.9}{0.2} && \entry{68.8}{0.4} & \entry{67.8}{1.2}\\
			\keyFieldL{64}  && \ent{0.0014\%} && \entry{51.6}{0.6} & \entry{48.3}{2.8} && \entry{87.5}{0.7} & \entry{88.1}{0.7} && \entry{86.5}{0.2} & \entry{86.5}{0.2} && \entry{63.5}{0.3} & \entry{64.8}{1.5}\\
			\keyFieldL{16}   && \ent{0.0004\%} && \entry{47.0}{1.4} & \entry{40.8}{4.5} && \entry{85.3}{0.3} & \entry{85.0}{0.2} && \entry{85.3}{0.1} & \entry{84.7}{0.1} && \entry{64.1}{0.8} & \entry{63.1}{1.1}\\
			\keyFieldL{4}   && \ent{0.0001\%} && \entry{39.5}{0.8} & \entry{36.7}{4.9} && \entry{83.5}{0.1} & \entry{82.6}{0.2} && \entry{83.8}{0.2} & \entry{80.0}{0.2} && \entry{61.4}{0.5} & \entry{59.3}{0.7}\\
			\bottomrule
		\end{tabular}}
	\caption{The performance of the fine-tuned BERT with $n$ trainable parameters in every \layernorm\ module on four different target tasks. Selecting the $n$ parameters from the outliers leads to better performance in most cases, compared to the random selection. For $n=256$, the results of outlier tuning are comparable with the \fullfinetune\ scenario.}
	\label{tab:outlier}
\end{table*}

			Outliers are high-magnitude weights in \layernorms\ appearing early in the pre-training process \cite{kovaleva-etal-2021-bert}. Transformer-based models perform significantly worse on downstream tasks when their outliers are disabled after the fine-tuning process \cite{kovaleva-etal-2021-bert}.

			In this experiment, we choose outliers as the set of $n$ weights whose values are farthest from the mean. Except for the outliers, all the parameters are frozen during fine-tuning. It should be considered that the specific dimensions where the outliers appear may not necessarily be the same across different layers.

			Table \ref{tab:outlier} presents the performance of the fine-tuned BERT in two different settings and for four different values of $n$: 4, 16, 64, 256. We also report the results for the corresponding sets of $n$ randomly selected weights. As can be observed, outliers tuning leads to competitive performance on most target tasks, despite using less than $0.0056\%$ of all the model parameters. Interestingly, tuning in the extremely constrained setting of $n=4$ still outperforms the frozen model, sometimes by significant margins (e.g., on \sts). Setting $n$ to higher values gives the model more capacity, bringing about higher performance.

			Overall, we can conclude that the high-magnitude weights in \layernorms\ play an important role in the effectiveness of these modules in parameter-efficient fine-tuning.

	\section{Conclusions}
		In this work, we study the efficiency of different modules in the transformer block of BERT to transfer knowledge from the pre-trained model to various downstream tasks. Our experimental results demonstrate that, contrary to what was suggested by previous work, every module can be a \textit{winning ticket}, achieving comparable performance to the full fine-tuning scenario. Among all modules, \layernorms\ prove to be the most reliable for transferability with a limited number of trainable weights, such that tuning them in only one layer can be sufficient for attaining performance on a par with that of the full fine-tuning. We find that the weights in these modules have notably high magnitudes compared to other modules, which could be the reason for their effectiveness. We examine this hypothesis through Outlier Tuning (tuning only the $n$ weights in each \layernorm\ whose values are farthest from the mean), limiting the number of tunable parameters to a significantly small fraction.

		Our results pave the way for better parameter-efficient fine-tuning of large language models without the need for costly algorithms to determine the optimum sub-network or introduce additional parameters for knowledge transfer.

	\section{Acknowledgment}
		We thank the anonymous reviewers for the constructive comments and suggestions that helped improve the paper. Sara Rajaee is funded in part by the Netherlands Organization for Scientific Research (NWO) under project number VI.C.192.080.

	\section{Limitations}
		We were subject to the constraints of computational resources; as a consequence, we reported results only for bert-base and chose the four smallest tasks of the \glue\ benchmark in the tuning single norms ({Section \ref{sec:sing}}) as well as outlier tuning (Section \ref{sec:outlier}). Obviously, the more trainable parameters a model has, the more accurate its results will be. Since our outlier tuning technique fine-tunes just a tiny portion of parameters, less than $0.006\%$ of the model weights, there is an upper bound on its learning capability.

	\bibliography{references}
	\bibliographystyle{acl_natbib}
\end{document}